\documentclass[journal,english]{IEEEtran}

\usepackage{times}
\usepackage{graphicx}
\usepackage{amsmath}
\usepackage{amssymb}
\usepackage{tabularx}
\usepackage{babel}
\usepackage{threeparttable}
\usepackage{subfigure}
\usepackage{color}

\usepackage[breaklinks=true,bookmarks=false]{hyperref}

\begin{document}

\title{
\begin{huge}
Pornographic Image Recognition \\ via Weighted Multiple Instance Learning
\end{huge}
}

\author{Xin Jin \quad Yuhui Wang \quad Xiaoyang Tan*\\
\thanks{Xin Jin, Yuhui Wang, and Xiaoyang Tan are all from the Department of Computer Science and Technology \& the Collaborative Innovation Center of Novel Software Technology and Industrialization, Nanjing University of Aeronautics and Astronautics, Nanjing 210016, China. Some of this work is done when Xin Jin works in Megvii Inc. (Face++).
{\tt\small \{x.jin, y.wang, x.tan\}@nuaa.edu.cn}}
\thanks{*Corresponding author: Tel.: +86-25-8489-6490/6491 (Ext 12106 ) (O); Fax:  +86-25-8489-2452; E-mail:x.tan@nuaa.edu.cn.}
}

\maketitle
\begin{abstract}
In the era of Internet, recognizing pornographic images is of great significance for protecting children's physical and mental health. However, this task is very challenging as the key pornographic contents (e.g., breast, private part) in an image often lie in local regions of small size. In this paper, we model each image as a bag of regions, and follow a multiple instance learning (MIL) approach to train a generic region-based recognition model. Specifically, we take into account the regions' degree of pornography, and make three main contributions. Firstly, we show that based on very few annotations of the key pornographic contents in a training image, we can generate a bag of properly sized regions, among which the potential positive regions usually contain useful contexts that can aid recognition. Secondly, we present a simple quantitative measure of a region's degree of pornography, which can be used to weigh the importance of different regions in a positive image. Thirdly, we formulate the recognition task as a weighted MIL problem under the convolutional neural network framework, with a bag probability function introduced to combine the importance of different regions. Experiments on our newly-collected large scale dataset demonstrate the effectiveness of the proposed method, achieving an accuracy with 97.52\% TPR at 1\% FPR, tested on 100K pornographic images and 100K normal images.

\end{abstract}


\IEEEpeerreviewmaketitle

\section{Introduction}\label{sec:introduction}
\IEEEPARstart{T}{he} rapid development of Internet has brought us great convenience in daily life, while on the other hand, made us easily exposed on some objectionable contents, one typical example of which is the pornography \cite{ries2014survey}. Among others (e.g., videos and texts), images are arguably the most common carrier for pornography on Internet. These pornographic images, which are typically anarchic and even illegal to sell in the adult bookstores, however can easily and widely spread on Internet through pornographic web sites, social network applications, instance messaging tools, and even e-mail attachments. Children should be prevented from browsing such images without question, while grown-ups may also not want to be exposed on them, for example when at workplaces. All these have made recognizing and filtering out pornographic images a rising concern nowadays.

While considerable progress has been made in the last two decades \cite{fleck1996finding,shih2007adult,deselaers2008bag,lopes2009bag,jang2011adult,moustafa2015applying,nian2016pornographic}, recognizing pornographic images is still an open problem in computer vision. There are large intra-class variations in pornographic images, due to the changes in background, scale, scenario, and human pose. In particular, the key objectionable contents (e.g., breast, private part) that distinguish a pornographic image from normal images often lie in local regions of small size, while the cluttered background (e.g., normal body, bed, room, etc) may sometimes make up a large portion of the image. As a consequence, when sharing similar background and human pose, the pornographic images may look very similar to some non-pornographic (sexy) images. We note that to avoid copyright issues, we do not post any pornographic image in this paper.

Due to these characteristics, an effective recognition method should be able to discover the informative local regions in pornographic images. Following this principle, most of traditional algorithms are decomposed into two phrases: extracting regions of interest (ROI) from complex backgrounds, and calculating hybrid features from extracted ROIs for recognition \cite{zhu2004adaptive,zheng2006shape,shih2007adult,hu2007recognition,zaidan2014four,ries2014survey}. Since pornographic contents often come with skin exposure, skin detection techniques are commonly employed for ROI extraction \cite{ries2014survey}. However, because skin detection itself is a very challenging problem, ROI-based approaches typically have very limited generalization ability. To overcome this, Kim \emph{et al.} \cite{kim2009malicious} explicitly model the possible pornographic contents with a set of semantic features such as ``breast", ``belly" and ``bottom". However, accurately detecting these semantic features is very difficult in practice, due to their small patch support and large appearance variation.

In this paper, we propose to train a generic region-based recognition model, which does not specialize in certain body part (e.g., breast) but aims at more general and discriminative visual patterns, for example, the female's upper body with exposed breast, the lower body with exposed private part, the close-up of sexual behaviour, etc. In contrast to the semantic feature approach \cite{kim2009malicious}, our target regions augment private body parts or sexual acts with useful context, and thus can be more reliably recognized. Furthermore, these regions are not required to contain \emph{complete} private body part or sexual act, and some non-typical pornographic regions (e.g., a region containing only a part of breast) will also do. In addition, our target regions typically take relatively larger positions in an image, which allows us to identify a pornographic image quickly with very few evaluations.

However, training such a region-based recognition model is a non-trivial task. Standard supervised learning requires bounding box annotations for the pornographic regions of interest. However, unlike the well-defined semantic features \cite{kim2009malicious}, the annotation process of our target regions is easily influenced by subjectivity, as different people may have different understanding of useful context. Furthermore, different regions may have different degree of pornography - intuitively, a region containing a complete breast is more pornographic than the one containing a part of breast. In some extreme cases, it is even difficult to determine the label of some ambiguous regions, i.e., a region containing a very small part of breast.

To address these issues, we follow a weakly supervised multiple instance learning (MIL) approach, which relaxes the need for direct annotation of target regions using a semi-automatic bag generation strategy, and integrates the importance (the degree of pornography) of different regions in positive bags into a unified learning model. Specifically, we make three main contributions within this MIL framework.
\begin{itemize}
\item We show that based on very few annotations of the key pornographic contents \footnote{In this paper, we refer to the private body parts (e.g., breast and genital) and sexual behaviours as the \emph{key pornographic contents} in images, as they are the key factors for distinguishing a pornographic image from normal images.} in an image, we can generate a bag of properly sized regions, among which the potential positive regions usually combine the key pornographic contents (e.g., breast) with useful context (e.g., upper body) that can aid recognition.

\item We define a simple quantitative measure of an arbitrary regions's degree of pornography, according to its overlapping ratio with the annotated key pornographic regions. This can be naturally used to weigh the importance of different regions in a positive bag, since intuitively a region containing a complete breast should contribute larger to the bag probability than that containing a small part of breast.

\item We formulate the pornographic image recognition task as a deep weighted MIL problem, with a bag probability function introduced to combine the importance of different regions. The resulting region-based classifier is very robust and accurate, and can be conveniently used for testing by transforming the test image into a bag of multi-scale regions.
\end{itemize}

To validate the effectiveness of our approach, we have collected a large scale dataset from Internet consisting of 138,000 pornographic images and 205,000 normal images. This dataset covers almost all types of pornographic images on Internet, and intentionally includes many challenging non-pornographic images (i.e., bikini). Our method produces excellent performance on this dataset - it achieves an accuracy with 97.52\% True Positive Rate (TPR) at 1\% False Positive Rate (FPR) and 55 FPS with GPU, tested on 100K pornographic and 100K normal images. We also show excellent performance for cross-database experiment, by testing the pre-trained model (trained on our dataset) on NPDI benchmark dataset \cite{avila2011bossa}.

\section{Relative work}\label{sec_relativework}
\subsection{Pornographic image recognition}
There is a long history of nudity detection and pornographic image recognition in computer vision and perceptual psychology. Early works mainly focused on finding naked people in images based on a human structure model \cite{fleck1996finding,forsyth1996identifying,forsyth1999automatic}. However, these works require that the pornographic images have completely naked people and simple backgrounds, while pornographic images on Internet generally have wide variations in background, scale, scenario, and human pose. To recognize general pornographic images, current methods can be roughly classified as feature-based, region-based, and body part-based.

Feature-based approaches emphasize the extraction of features from the \emph{whole} image, and popular approaches include the bag-of-feature (BoF) approach \cite{deselaers2008bag,lopes2009bag,ulges2011automatic,sui2012research}, and the deep convolutional neural networks (CNNs) approach \cite{moustafa2015applying,nian2016pornographic}. The BoF approach encodes local image features (e.g., SIFT \cite{lowe2004distinctive}) with a visual vocabulary, and hence can capture some distinct local patterns of pornographic images. But due to the use of hand-crafted features, the BoF approach usually has limited discriminative power. By contrast, the CNN-based approach can automatically learn discriminative representations from large data. However, since they directly adapt the off-the-shelf CNN architecture to model the whole pornographic images, some crucial local details (e.g., breast) are largely ignored.

Region-based approaches extract features for recognition based on the detection of regions of interest (ROIs) in the image. Among others, skin regions are widely considered as the ROIs for pornography \cite{jiao2001detecting,zhu2004adaptive,zheng2006shape,shih2007adult}. To detect skin-like pixels, the input image is typically transformed from the RGB model to the RCbCr color space, and a pixel is considered to be skin-like when it satisfies a few linear constraints in RCbCr space \cite{garcia1999face}. Based on detected skin regions, multiple hand-crafted features (e.g., color, shape and texture features) are extracted for recognition. Compared to feature-based approaches, the region-based methods are more robust against background clutter, but run the risk of inaccurate ROI detection since skin detection is a challenging problem in itself.

Body part-based approaches define several pornography-related semantic features such as ``breast", ``belly" and ``bottom", and train corresponding body part detectors for these features \cite{kim2009malicious,lv2011pornographic}. Given an image, these body part detectors are used to scan the image, and the detection results are then arranged in a semantic feature vector for classification. However, these body part detectors are plagued by the problem of ambiguity and very likely to generate false positive detection, due to their small patch support and large appearance variations in training.

In this paper, we propose to train a generic region-based model for pornographic image recognition. Unlike traditional region-based approaches, our approach does not rely on ROI detection, and directly identify pornographic images by evaluating our model on the test image at different positions and scales. Furthermore, while similar in spirit with the body part-based approaches  \cite{kim2009malicious,lv2011pornographic}, our generic recognition model aims at more general and discriminative pornographic contents in images, which typically combine useful context with the key pornographic contents and thus can be more reliably recognized than the individual private body parts.

\subsection{Multiple Instance Learning}
Research on Multiple Instance Learning (MIL) studies the problem where a real-world object is associated with a single class label but described by a number of instances. The MIL framework was first formalized by Dietterich et al. \cite{dietterich1997solving} to investigate drug activity prediction, and then have been applied to diverse applications including image classification \cite{chen2004image,chen2006miles,xiao2014similarity}, face detection \cite{zhang2005multiple,zhang2008multiple}, image annotation \cite{tang2010image,hong2014image}, and saliency detection \cite{wang2013saliency}. Since space does not allow for a full review, we will only focus of MIL algorithms most related to ours.

MIL has been widely used for object recognition and localization, by considering each image as a ``bag" of examples given by tentative object windows, and assuming that positive images contain at least one positive instance object window, while negative images only contain negative windows \cite{cinbis2014multi}. Traditional MIL algorithms assume that all of the instances contribute equally and independently to the bag's label. But, sometimes, it's desirable to take into account the instance importance in the learning procedure. For example, Zhang \emph{et al.} \cite{zhang2013real} proposed an online weighted MIL algorithm for visual tracking. They defined a novel bag probability function that combines the weighted instance probability, leading to a more robust and much faster tracker. Our approach is also based on weighted MIL, but differs from \cite{zhang2013real} in bag generation, weight definition, and most importantly, while \cite{zhang2013real} performed an online MIL boosting technique for feature selection, we formulate the weighted MIL algorithm under the CNN framework to learn discriminative features.

To the best of our knowledge, Li \emph{et al} \cite{li2015pornographic} is the only one to employ MIL technique for pornographic image recognition. They treat each image as a bag, and low-level visual features of divided sub-blocks as instances. Specifically, they first perform spatial pyramid partition to divide an image into blocks, and then extract three low-level visual features (i.e., color features, LBP \cite{ojala2002multiresolution}, and HOG \cite{dalal2005histograms}) on these blocks, resulting a representation consisting of three different multi-instance bags. However, each bag in \cite{li2015pornographic} is then transformed into a single representation vector (metadata), and a standard single instance learning method is used to solved the MIL problem. By contrast, we treat each image as a bag of regions throughout the algorithm, and learn a region-based classifier that can be use not only to identify pornographic images but also to localize the rough pornographic regions in these images.


\section{The Approach}\label{sec_methodology}
As mentioned before, the key pornographic contents in an image often lie in local regions, and the background regions may be distractive for recognition. As a consequence, recognition methods based on the whole image are very sensitive to background clutter. Considering this, we propose to model each image as a bag of regions, and follow a multiple instance learning (MIL) approach to train a generic region-based recognition model.

In this section, we first describe our bag generation strategy, which is based on the annotations of the key pornographic contents in images during training. Based on these annotations, we further present a quantitative measure of the regions' degree of pornography in positive bags. With the region-based representation and pornography measure, we formulate the recognition task as a weighted MIL problem under the convolutional neural network framework.

\subsection{Bag Generation} \label{subsec:bag_generation}

Bag generation is crucial for a MIL algorithm \cite{zhou2004multi}. In this section, we mainly focus on the bag generation strategy in training, while the testing process will be detailed in Section \ref{subsec:testing}. Currently, the most popular approach is to consider each image as a ``bag" of examples given by region proposals \cite{uijlings2013selective,zitnick2014edge}. However, as mentioned before, our target regions are the general pornographic regions that combine pornographic contents with useful contexts while excluding distractive backgrounds. These regions (e.g., the female's upper body with exposed breasts) typically cannot be regarded as ``objects", and thus are not in accordance with the goal of region proposal algorithms.

Besides, we argue that a good positive bag in training should strive to cover various regions that may encounter in testing, including the typical pornographic regions, the non-typical positive regions and the negative (non-pornographic) regions. The non-typical positive regions, such as the upper body containing only a part of breast, are very important since we cannot always obtain typical pornographic regions for a test image. The negative regions in positive images are also very useful, since they are by nature difficult training instances. In addition, scale normalization is important for model training, and regions from different positive bags should be of similar scale. To satisfy these requirements, we propose to annotate the key pornographic contents in positive images first, and then use these annotations to guide the bag generation process.

There are two main types of the key pornographic contents, i.e., the private body parts, and sexual behaviours.We perform bounding box annotation for these key pornographic contents in the training set, and for the sexual behaviour, we just annotate the most compact regions that are sufficient to characterize these behaviours. In general, there are only a small number of annotations (less than 10) required for a pornographic image, and the annotation process is unlikely to be affected by subjectivity due to the clear definition of key pornographic contents.

With these annotations, we can obtain a variety of regions for a positive image by moving a window around each annotated pornographic region. In particular, the window size is adaptively computed according to size of each annotated pornographic region, and the aspect ratio is the same to that of the image \footnote{In both training and testing phases, the regions are resized to 224$\times$224 to satisfy the requirement of GoogLeNet model \cite{szegedy2014going}}. In our current system, we randomly set the window size to be 2, or 2.5, or 3 times of the annotated region. Here the slight scale change is introduced to improve model's robustness against scale variation, and these three similar scales are selected by cross validation, and thus are experimentally proved to be able to include useful context information.

By randomly moving the windows from each annotated region with different displacements, we can generate about 100 regions for a positive image, among which some are typical positive regions, some are non-typical positive regions, and some are non-pornographic regions. For negative images, the bag generation process is the same with the test procedure, as described in \ref{subsec:testing}.

\subsection{Pornography Measurement}

Intuitively, the typical positive regions should be more pornographic than the non-typical ones. That is, a region containing a complete breast should contribute larger to the bag probability than the one containing only a small part of breast. We take this into account, and present a simple quantitative measure of an arbitrary region's degree of pornography.

Our measurement is based on the overlapping ratio between the generated regions and the annotated key pornographic regions. We set a region's degree of pornography to be a scalar number between 0 to 1, by finding the maximum overlapping ratio between this region and the annotated key pornographic regions. Furthermore, by normalizing these degrees of pornography (i.e., sum to 1), we can obtain the weights (importance) of different regions in a pornographic image. In the following, we will describe how to integrate the importance of different regions into a deep learning framework.

\subsection{Deep Weighted MIL}
Inspired by the great success of convolutional neural network (CNN) in image classification \cite{krizhevsky2012imagenet,szegedy2014going}, in this section, we formulate the recognition task as a weighted Multiple Instance Learning (MIL) problem under the CNN framework. In particular, we introduce a bag probability function to combine the importance of different region in positive images.

In our MIL setting, each image $\mathbf{X}$ is modeled as a bag of $n$ regions $\{\mathbf{x}_i|i\! =\!1,...,n\}$. Given a region, the deep CNN extracts layer-wise representations from the first convolutional layer to the last fully connected layer. Our CNN architecture is inspired by the GoogLeNet model \cite{szegedy2014going}. The output of the last fully connected layer is a 1,000 dimensional vector, followed by a softmax layer to transform it into a probability distribution for objects of 1,000 categories. As pornographic image recognition is a two-class classification problem, we re-design the output of the last fully connected layer to be a 1 dimensional vector, and transform it into a Bernoulli distribution via the sigmoid function. Formally, given region $\mathbf{x}_i$, we denote the output of the last fully connected layer as $h_i$, and let $p_{i}^{+}$ and $p_{i}^{-}$ be the probability that region $\mathbf{x}_i$ is pornographic or non-pornographic respectively. Then, we have
\begin{equation}\label{eq:instance_prob}
\begin{split}
p_{i}^{+}&=\frac {e^{h_i}} {e^{h_i}+1}, \\
p_{i}^{-}&=\frac {1} {e^{h_i}+1}.
\end{split}
\end{equation}

Now, let's take into account of the importance of different regions in a positive bag, and aggregate the instance probabilities \ref{eq:instance_prob} into a positive bag probability. As aforementioned, we can obtain the weight $w_i$ for each region $\mathbf{x}_i$ by normalizing the degree of pornography for all extracted regions. Intuitively, we can directly assign these weights to the instance probabilities \ref{eq:instance_prob}, and combine them to form the bag probability for a positive bag. However, we notice that the weights of some regions in a positive bag are just 0, i.e., these regions have no overlapping with the annotated key pornographic regions, and hence should be treated as negative instances. Considering this, we divide each positive bag into two sub-bags, i.e., a positive sub-bag where the regions are all assigned with positive weights, and a negative sub-bag where the weights are all 0. Let $n^{+}$ and $n^{-}$ be the number of the regions in the positive sub-bag and negative sub-bag respectively, we can define the sub-bag probabilities as follows,
\begin{equation}\label{eq:subbag_prob}
\begin{split}
p^{+} &= \sum_{i=1}^{n^+} w_{i}p_{i}^{+}, \\
p^{-} &= \frac{1}{n^{-}}\sum_{i=1}^{n^-} p_{i}^{-},
\end{split}
\end{equation}
where the probability $p^{-}$ for the negative sub-bag is computed by just averaging the instance probabilities.
With the sub-bag probabilities \ref{eq:subbag_prob}, we simply define the positive bag probability function as
\begin{equation}\label{eq:bag_prob_fun}
  p = p^{+}p{-}.
\end{equation}

Then, we reach the following loss function (negative bag log-likelihood function) for a positive image:
\begin{equation} \label{eq:loss_fun}
L = -1\{n^+>0\}\log p^{+} -1\{n^->0\}\log p^{-}.
\end{equation}
Actually, we notice that the bag probability function and the loss function for a negative image can also be written as
 \ref{eq:bag_prob_fun} and \ref{eq:loss_fun}, since we always have $n^{+} = 0$ and $n^{-} > 0$ for a negative image. The gradient for \ref{eq:loss_fun} is calculated via back propagation,
\begin{equation}
\frac {\partial L} {\partial h_i}=
\big(-1\{w_i > 0\} \frac 1 {p^+}w_i + 1\{w_i = 0\}\frac 1 {p^-n^-}\big) p_{1,i}p_{0,i}.
\end{equation}

We resize each region to 224$\times$224, and use the open-source package Caffe \cite{jia2014caffe} to extract deep features based on the GoogLeNet model \cite{szegedy2014going}. We re-define the objective of GoogLeNet as \ref{eq:loss_fun}, and perform domain-specific fine-tuning to train our region-based recognition model. In the next section, we will turn to the testing process of the learned model.

\subsection{Multi-scale Testing} \label{subsec:testing}

In order to reduce the appearance variations in training, the regions extract from pornographic images are normalized to similar scales according to our annotations. However, in the testing phase, we may encounter pornographic images of various scales. To address this issue, we adopt a simple multi-scale detection window approach. Specifically, we extract regions at 3 scales for a test image. The smaller two scales are set to 1/2 and 2/3 of the width and height of the whole image, while the large scale is set to the whole image. For the smaller two scales, we extract 5 regions from four corners and the center respectively. Therefore, we have 11 regions in total. We resize these regions into 224$\times$224, and if one region is recognized as pornographic, this image is considered as pornographic, while otherwise non-pornographic. In practice, we usually don't need to classify all these regions for a positive image, since once a region is recognized as pornographic, the test process is stopped.

\section{Experiments}\label{sec:experiments}
In this section, we present a comparative performance evaluation and discussion of the proposed approach. For this, we have collected a large scale dataset from Internet, which covers almost all kinds of pornographic images on Internet, with large variation of backgrounds, scenarios, lightings and poses. Our most significant results are on this dataset, including the comparisons to traditional methods, commercial systems, and deep learning-based baselines. We also report the running time performance of our method. Furthermore, we validate the effectiveness of the proposed components of the system, and discuss the advantages and limitations of our method.

Researchers have recently reported results on the NPDI pornography dataset \cite{avila2011bossa}. Since the NPDI dataset does not contain the annotations required to train our model, we opt to test on NPDI the model learned from our dataset to demonstrate its generalization capability cross dataset.

\subsection{Datasets}

\textbf{Our Dataset:} We have collected a large scale dataset consisting of 138,000 pornographic images and 205,000 normal images from Internet. In particular, we download pornographic images from four pornographic web sites using a web crawler software. The downloaded pornographic images are mixed with a small amount of normal images, and we exclude them manually. Our pornographic image set covers almost all kinds of pornographic images on Internet. Overall, we categorize them into three groups, namely, \emph{regular nudity}, \emph{sexual behaviour}, and \emph{unprofessional porn}. The \emph{regular nudity} refers to the images of nude people produced by professional photographer. The \emph{sexual behaviour} means images depicting sexual behaviour, which are also produced by professional photographer. The \emph{unprofessional porn} mean pornographic images taken by unprofessional photographer including both nudity and sexual behaviour. The unprofessional porn images usually contain complex backgrounds and the image quality is poor.

The normal images of our dataset are also downloaded from Internet, which can be further categorized into three groups, namely, \emph{scantily-clad people}, \emph{normal people} and \emph{no people}. In particular, the \emph{scantily-clad people} images include bikini, seductive images and man or baby with bare upper body, which are similar with pornographic images in appearance, and images of \emph{normal people} mean that people in these images are normally dressed, while the images of \emph{no people} consist of various normal images of natural scenery, animals and living goods and so on, containing no people.

From the pornographic images, we randomly select 33,000 images to annotate their key pornographic contents with bounding boxes. To collect, clean, and annotate these images, we employ six person and spend about two months. In our experiments, we use the annotated 33,000 pornographic images and 100,000 randomly selected normal images as the training set, randomly select 5,000 pornographic images and 5,000 normal images from the remaining images as the validation set, while the remaining 100,000 pornographic images and 100,000 normal images are used as the test set. Furthermore, to present detailed comparison on different types of images, we roughly split the test images into six groups, i.e., pornographic images: 19,207 \emph{regular nudity}, 37,018 sexual behaviour, 43,775 \emph{unprofessional porn}; and normal images: 68,873 \emph{scantily-clad people}, 20,230 normally \emph{dressed people} and 10,897 \emph{no people} images.

\textbf{NPDI dataset:} The NPDI pornography dataset \cite{avila2011bossa} contains nearly 80 hours of 400 pornographic and 400 non-pornographic videos. The videos are already segmented into shots, and on the average, there are almost 20 shots per video. In total, 16,727 key frames selected from the videos, 10,340 normal images and 6,387 pornographic images. However, we find that 1,198 of the 6,387 pornographic images are incorrectly labeled; hence we remove them from our experiment. Since the NPDI dataset does not have annotations of the key pornographic contents, we do not use it for model training, but for \emph{cross-dataset testing}.

\subsection{Experiment on Our Dataset}
\begin{table}[!htb]
\centering
\caption{Comparison of Detection Rate (\%)}\label{tab_exp_DR}
\scriptsize
\begin{threeparttable}
\begin{tabular*}{0.47\textwidth}{@{\extracolsep{\fill}}llll}
\hline
Methods & Porn & Normal & All\\
\hline
Region-of-Interest Method \cite{shih2007adult}  & 68.30 & 61.41 & 64.85\\
Improved RoI Method \cite{Yu2014Skin} & 76.53 & 66.78 & 71.65\\
Bag-of-Feature Method \cite{lopes2009bag}  & 79.79 & 71.87 & 75.83\\
Improved BoF Method \cite{Zhang2013An} & 73.76 & 82.16 & 77.96\\
\hline
Commercial System 1 \cite{qcloud}& 53.04 & \textbf{99.16} & 76.11\\
Commercial System 2 \cite{yahoo}& 84.37 & 81.83 & 83.10\\
\hline
Deep Image-based CNN \cite{nian2016pornographic} & 88.18 & 93.66 & 90.92\\
Deep Fused CNN \cite{moustafa2015applying} & 92.67 & 89.22 & 90.94 \\
Deep Region-based CNN & 95.65 & 95.04 & 95.35\\
Deep Part Detector  & 92.27 & 52.49 & 72.36\\
Deep MIL & 94.15 & 95.84 & 94.97 \\
\hline
\textbf{Ours} & \textbf{98.18} & 98.51 & \textbf{98.35}\\
\hline
\end{tabular*}
\end{threeparttable}
\end{table}

\begin{figure}[!htb]
\centering
\includegraphics[width=0.45\textwidth]{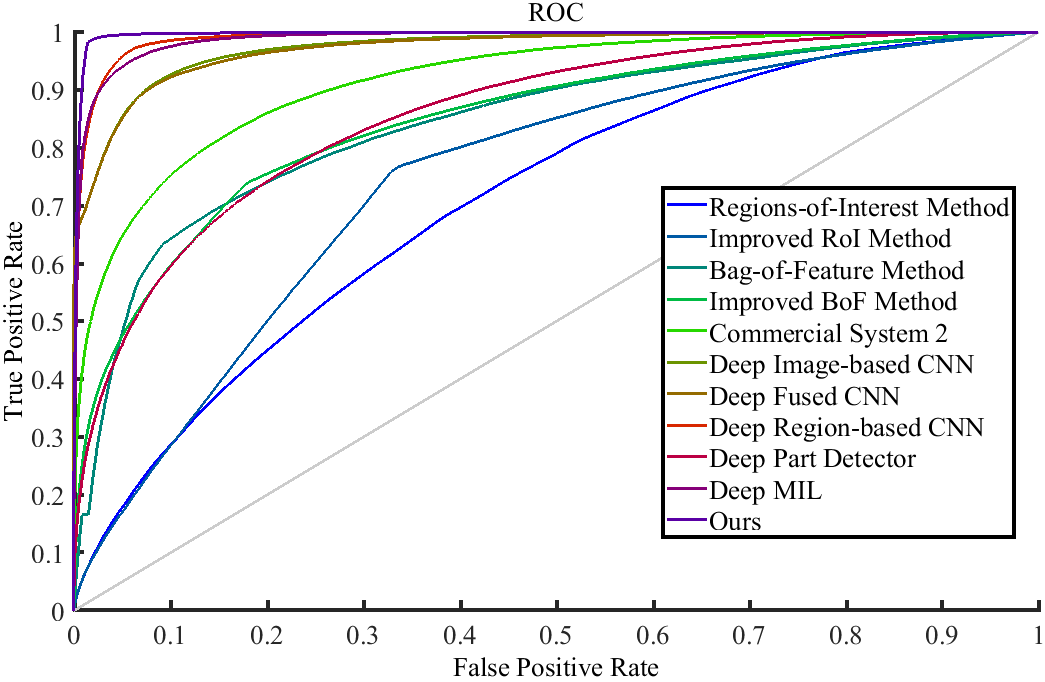}
\caption{ROC curve of different methods for pornographic image recognition. Our method significantly outperforms traditional methods and shows some advantage over the in-house baselines using deep representation. In particular, we achieve accuracy of 97.52\% TPR at 1\% FPR.}
\label{fig_ROC}
\end{figure}

On our newly-collected large scale dataset, we first compare our method with four traditional methods using shallow low-level features and two commercial systems using deep learning techniques, then we compare it with another five deep learning baselines implemented by ourselves. We use the detection rate and the ROC curve to measure performance. The detection rate is computed by the fraction of identified images out of the pornographic/normal images set/subset. Besides comparison to other methods, we also report our running time performance, and validate the proposed system components, and present some discussions about our method.

\subsubsection{Comparison to Traditional Methods}
Traditional works on pornographic image recognition mainly focus on two methodologies: Regions of Interest (ROI) based methods \cite{zhu2004adaptive,zheng2006shape,shih2007adult,Yu2014Skin} and Bag-of-Feature based methods \cite{deselaers2008bag,lopes2009bag,ulges2011automatic,Zhang2013An}. However, to best of our knowledge, there is no off-the-shelf recognition systems available for comparison. Hence, we implement two representative ROI-based methods and two Bag-of-Feature methods respectively. They are the retrieval-based method that extracts color, shape and skin features on ROI obtained by skin detection \cite{shih2007adult} and its improved version with a novel skin detection technique \cite{Yu2014Skin}, and the Bag-of-Feature (BoF) approach based on Hue-SIFT descriptor \cite{lopes2009bag} and its improved version based on visual attention model \cite{Zhang2013An}. Both of these methods have reported good performances on their small scale datasets.

Rows 1 to 4 and last row in Tab. \ref{tab_exp_DR} show the detection rate of the traditional methods and ours on pornographic images, normal images and all test set. Fig. \ref{fig_ROC} shows the ROC curve. We can observe that the proposed method significantly outperform the ROI-based method \cite{shih2007adult} and its improved version \cite{Yu2014Skin}, as well as the Bag-of-Feature (BoF) method \cite{lopes2009bag} and its improved version \cite{Zhang2013An}. The results on different types of pornographic images and normal images are shown in Fig. \ref{fig_hist} In particular, on the \emph{scantily-clad people} subset of normal images, our detection rate is almost twice of the ROI-based method \cite{shih2007adult}.

\subsubsection{Comparison to Commercial Systems}
As pornographic image recognition is very important and useful in practice, some commercial systems have been developed and shared in recently years. We choose two representative commercial systems, which use deep learning techniques and are launched in web pages with URLs \cite{qcloud,yahoo}. We refer to them as \emph{Commercial System 1} and \emph{Commercial System 2} respectively. We note that since the these two systems are trained with their own data, a completely objective comparison between them and our system is not possible.

The performance of these two commercial systems on our test set in shown in Row 5 and 6 of Tab. \ref{tab_exp_DR}. The overall performance of these two commercial systems is inferior to that of our method. But we can see that commercial system 1 achieves the best accuracy (99.16\%) on our normal image set, comparing to all other methods. In contrast to its high performance on normal image set, it has rather poor prediction accuracy (53.04\%) on our pornographic image test set. The commercial system 2 achieves 83.10\% accuracy on our test set, which is higher than two traditional algorithms but still have a clear gap with our method.

\subsubsection{Comparison to Deep CNN Baselines}
To the best of our knowledge, there are two deep CNN-based methods for pornographic image recognition \cite{moustafa2015applying,nian2016pornographic}, which however have not released their code and training data. To conduct a comprehensive comparison with deep CNN-based baselines, we implement the algorithms described in \cite{moustafa2015applying,nian2016pornographic} and other three CNN-based variations with our training data.

First, since the main idea of \cite{nian2016pornographic} is to directly fine-tune the off-the-shelf CNN architecture with their own data, we implement a similar deep CNN version by fine-tuning the GoogLeNet \cite{szegedy2014going} with our training data, and refer it as \emph{deep image-based CNN}. Second, we re-implement the algorithm in \cite{moustafa2015applying} which fuses the recognition results by AlexNet \cite{krizhevsky2012imagenet} and GoogLeNet \cite{szegedy2014going}, and refer it as \emph{deep fused CNN}. Third, we design a naive version of region-based CNN (similar to \cite{girshick2014rich}) using our bag generation strategy. We assume that if a region's degree of pornography is larger than 0.5, it is labeled as positive and otherwise negative. In this setting, we can train a region-based recognition model using standard supervised learning. We refer to this method as \emph{deep region-based CNN}. Fourth, based on our annotations, we also train three types of body part detectors with GoogLeNet, including female breast, female sexual organ and male sexual organ detectors. We resize the annotations for these body parts into 224$\times$224 patches, and fine-tine GoogLeNet with these patches. \footnote{The accuracies are 95.02\%, 92.72\%, 94.27\% for female breast, female sexual organ and male sexual organ detectors on validation set of their positive and negative patches, respectively.} The trained body part detectors are then used for testing by scanning the test image in a cascaded manner and considering the image as pornographic when one detector has positive response. We refer to this method as \emph{deep part detector}. Finally, we use the standard MIL paradigm to train a region-based detector, without using the weighting strategy. We refer to this method as \emph{deep MIL}.

Rows 7 to 11 in Tab. \ref{tab_exp_DR} show the detection rate of these deep CNN-based methods, and Fig. \ref{fig_ROC} shows the ROC curve. Our method consistently outperforms all deep CNN-based baselines, although most of them can outperform the traditional methods by a large margin. In particular, the deep image-based CNN method outperforms two commercial systems that also employ deep learning techniques. We conjecture that this may be caused by the difference of training data. Furthermore, the deep region-based CNN achieves a significant improvement over the deep image-based CNN method, which confirms the effectiveness of region-based modeling strategy. In addition, we observe that the body part detector method performs very poorly on the scantily-clad images as shown in Fig. \ref{fig_hist}. This result is consistent with our intuition since these images only have small cover on private parts especially on female breast, which may lead to false positive detection.

\subsubsection{Cross-validation Performance}
In this section, we verify the generalization ability of our method via cross-validation. In particular, we perform 10-fold cross-validation on the training data. Tab. \ref{tab:exp:cross-validation} gives the validation performance of each round as well as the average accuracy. Our algorithm obtains consistent results among different rounds, and achieves good average performance. With 90$\%$ training data, the average cross-validation accuracy is 97.32$\%$, which is very close to the test accuracy with all training data.

Besides cross-validation, we perform two extra experiments to further verify the stability of our algorithm, by changing the setting of training while keeping the test set unchanged. In particular, we first train our model 10 times with stochastic gradient descent(SGD), and then test these models on the test set respectively. We surprisingly find that all these models almost produce the same result as reported in Tab. \ref{tab_exp_DR}. Furthermore, we also test the 10 cross-validation models on the full test set, which are trained with 90$\%$ training data. The result is shown in Tab. \ref{tab:exp:multi-test}. We see that these models show consistency in testing, and the accuracy only drops a little when using the reduced training data.

\begin{table}[!htb]
\centering
\caption{Cross-validation Performance(\%)}\label{tab:exp:cross-validation}
\scriptsize
\begin{threeparttable}
\begin{tabular*}{0.47\textwidth}{@{\extracolsep{\fill}}llll}
\hline
Round & Porn & Normal & All\\
\hline
1 &96.68 &94.84 &95.29 \\
2 &97.44 &98.34 &98.12 \\
3 &97.66 &97.34 &97.42 \\
4 &98.18 &95.84 &96.42 \\
5 &96.64 &96.32 &96.40 \\
6 &94.92 &95.84 &95.61 \\
7 &98.04 &99.96 &99.49 \\
8 &97.56 &95.36 &95.90 \\
9 &96.04 &99.98 &99.26 \\
10 &97.58 &99.78 &99.24 \\
\hline
Mean &97.07$\pm$0.96 &97.40$\pm$1.96 &97.32$\pm$1.54\\
\hline
\end{tabular*}
\end{threeparttable}
\end{table}

\begin{table}[!htb]
\centering
\caption{Multiple Test Performance with Reduced Training Data\%)}\label{tab:exp:multi-test}
\scriptsize
\begin{threeparttable}
\begin{tabular*}{0.47\textwidth}{@{\extracolsep{\fill}}llll}
\hline
Round & Porn & Normal & All\\
\hline
1 &96.67 &95.76	&96.22\\
2 &96.86 &96.41 &96.63\\
3 &97.42 &95.28	&96.35\\
4 &93.64 &93.10	&93.37\\
5 &96.80 &94.12 &95.46\\
6 &93.64 &93.10 &93.37\\
7 &97.94 &97.78 &97.86\\
8 &97.22 &93.40	&95.31\\
9 &96.24 &98.18 &97.21\\
10 &97384 &97.90 &97.64\\
\hline
Mean & 96.38$\pm$1.44 & 95.50$\pm$1.92 & 95.94$\pm$1.51\\
\hline
\end{tabular*}
\end{threeparttable}
\end{table}

\subsubsection{Testing on Gray Images}
Almost all the training and testing images in our experiments are color images. But in practice, we may also encounter gray pornographic images. Therefore, we also test our model trained with color images on the black and white copy of our test set. The detection rate on positive images, negative images and all images are 82.23\%, 99.09\%, and 90.66\% respectively. The test result is promising, since we did not include any gray images during training. In particular, the test result on gray negative images is even better than the counterpart on original color negative images.

\subsubsection{Running Time Performance}
We implement our system with the open-source package Caffe \cite{jia2014caffe}, and measure it on an Intel Xeon E5-2630 CPU (2.60 GHz, 12 core) and a GeForce GTX Titan X GPU respectively. We achieve 5 FPS with CPU, and 55 FPS with GPU. That is, the proposed method is highly efficient and real-time capable when implemented with GPU.

\subsubsection{Validation of System Components}

\begin{figure}[!htb]
\centering
\includegraphics[width=0.48\textwidth]{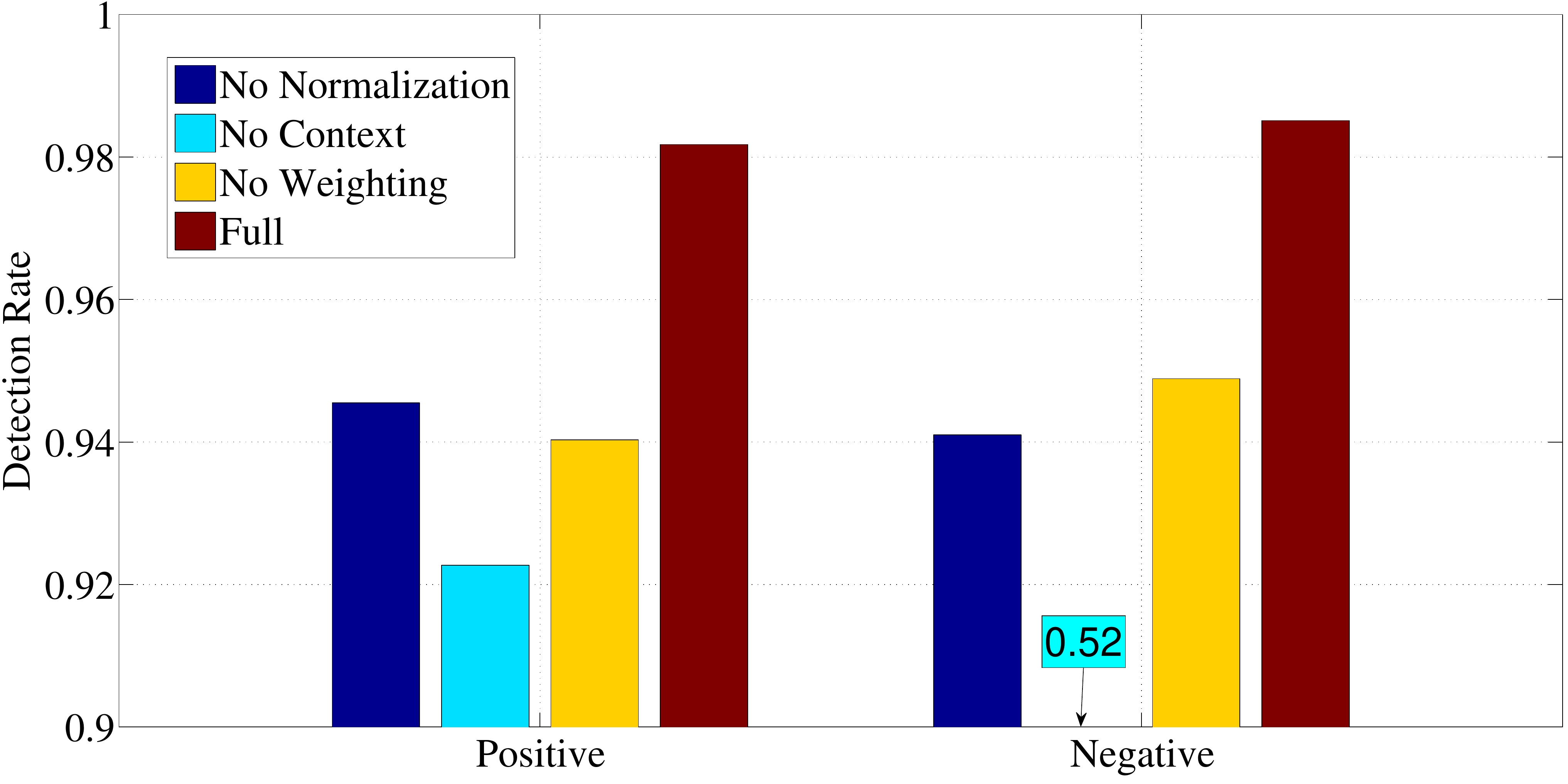}
\caption{The contribution of the components to the performance, including the scale normalization for training regions, the effect of useful context, and the weighting strategy for MIL.}
\label{fig_component_validation_hist}
\end{figure}

In this section, we validate the effectiveness of three proposed components of our system, including the normalization for training regions, the inclusion of context information, and the weighting strategy for MIL. To investigate the contribution of the individual component, we conduct a series of experiments on our dataset by removing each of the components in turn while leaving the remaining components in place. Fig. \ref{fig_component_validation_hist} gives the results. In particular, the pipeline that removes useful context can be seen as generic private body part detector, which achieves very poor performance on the negative images. This suggests that the human body parts usually lack distinct patterns without the cue from useful context. We also observe that scale normalization according to our annotations can boost the performance of our method, since it ensures that the positive regions are dominated by pornographic contents to some extent. Lastly, our method clearly benefits from the weighting strategy of MIL, which confirms the importance of incorporating the region's degree of pornography into MIL.

\subsubsection{Discussions}
\begin{figure*}[!htb]
\centerline{
\subfigure[Pornographic Image Results]{\includegraphics[width=0.5\textwidth]{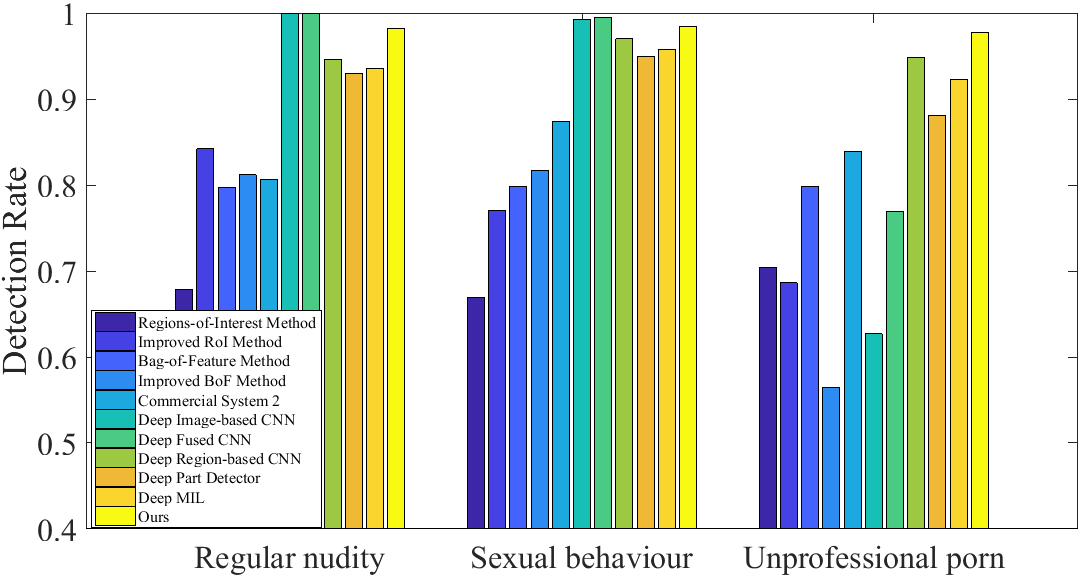}
\label{fig_pos_hist}}
\subfigure[Normal Image Results]{\includegraphics[width=0.5\textwidth]{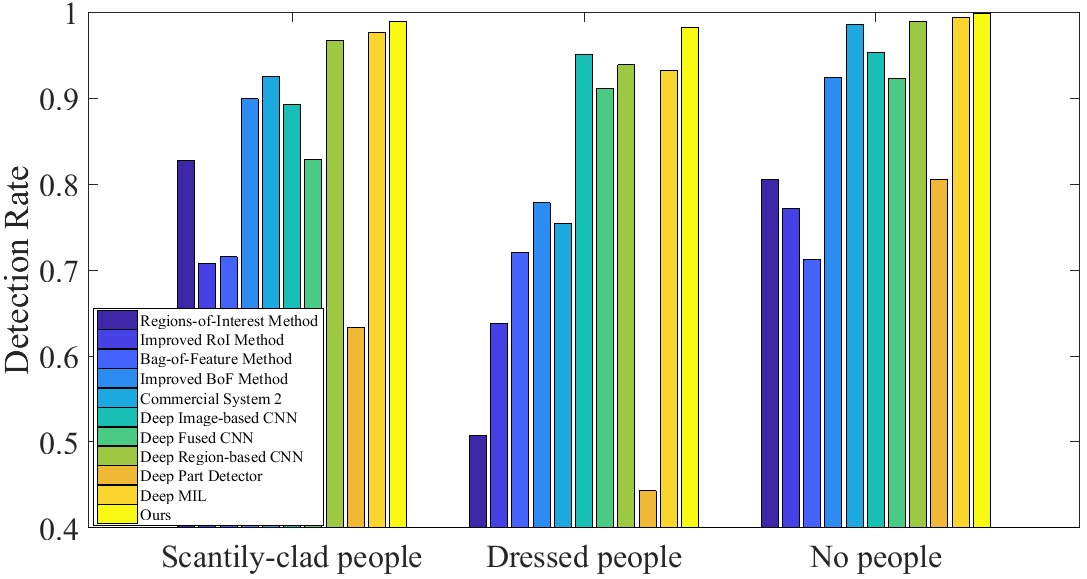}
\label{fig_neg_hist}}
}
\caption{Illustration of results (detection rate) on different types of pornographic images and normal images.}
\label{fig_hist}
\end{figure*}
In this section, we present discussions about the advantages and limitations of our method, by analyzing our performance on different types of images.

We summarize the performance of our method and other comparative methods on different types of test images in Fig. \ref{fig_hist}. Our method performs extremely consistently on different images, while other methods rise and fall greatly. In particular, our method shows obvious advantages over all other methods on the \emph{unprofessional porn} images. This part of pornographic images mainly come from unprofessional photographs in daily life, which exhibit large variations in background, scale, lighting and people pose. This suggests that our method can capture the essential pornographic contents of these images to some extent, while other methods (e.g., deep image-based CNN) may be effected by the large background of these images. We notice that our performance on scantily-clad images are even better than that on normally dressed images. This seems a little strange, and we conjecture that it's because the deep representations based on useful context information have the capability to discriminate exposed private body parts and private body parts with a small cover, while the normally dressed images in life have larger variations in clothe styles, poses and scenes compared to scantily-clad images, which may bring in challenges for recognition.

Many of our true positive images exhibit large variations of lighting, pose, scale and occlusion. On the other hand, we note that false detections are likely to happen in the local regions that look like exposed female breast, while actually they are exposed male breasts. In addition, since we consider the exposure of private body parts or sex acts as the criteria for pornographic images, our current system cannot properly distinguish some art works with exposed private part from pornographic images. But fortunately, such nude art works are rather a small part of Internet images.

Furthermore, we also find many challenging true negative images on our dataset, including seductive images, bikini girls, men with naked upper body, and naked babies. These images, which typically show similar appearance with pornographic images, can to some extent validate the capability of our method to distinguish pornography from some related concept such as seduction and nudity. In addition, the typical false negative images of our method often show very little private body parts such that they can hardly be correctly detected.

\subsection{Cross-database Experiment on NPDI Dataset}
\begin{table}[!htb]
\centering
\caption{Comparison on NPDI by Mean Average Precision (MAP)}\label{tab:exp_NPDI_MAP}
\scriptsize
\begin{threeparttable}
\begin{tabular*}{0.4\textwidth}{@{\extracolsep{\fill}}llll}
\hline
Methods & MAP\\
\hline
Region-of-Interest Method \cite{shih2007adult}  & 0.712\\
Improved RoI Method \cite{Yu2014Skin} & 0.756\\
Bag-of-Feature Method \cite{lopes2009bag}  & 0.811 \\
Improved BoF Method \cite{Zhang2013An} & 0.823\\
Deep Image-based CNN \cite{nian2016pornographic} & 0.874\\
Deep Fused CNN \cite{moustafa2015applying} & 0.919 \\
Deep Region-based CNN & 0.878\\
Deep Part Detector  & 0.870\\
Deep MIL & 0.860\\
BossaNova \cite{avila2013pooling} & 0.964\\
\hline
\textbf{Ours} & \textbf{0.975}\\
\hline
\end{tabular*}
\end{threeparttable}
\end{table}

On the NPDI Pornography dataset, we test our recognition model trained on our dataset to demonstrate it's generalization capability across datasets. The classification performance is evaluated using the standard metric for this dataset, the Mean Average Precision (MAP). Our model achieves good performance, reaching a MAP of 0.975, while the best performance in literature on this database is 0.964 (MAP) \cite{avila2013pooling}.

Table \ref{tab:exp_NPDI_MAP} shows detailed comparison between our method and many other methods which are also trained on our dataset. Our method consistently outperforms four traditional methods and five deep learning baselines. We note that our model performs well, although there are some differences between the styles of the training pornographic images from our database and the test pornographic images from NPDI dataset.

\section{Conclusion}\label{sec_conclusion}
In this paper, we follow a multiple instance learning (MIL) approach to address the problem of pornographic image recognition. In particular, we exploit the annotations of the key pornographic contents in the training phase. These annotations help us to generate a variety of regions with different degree of pornography, and to define a quantitative measure of these regions' degree of pornography. Based on these, we formulate the recognition task as a weighted MIL problem under the CNN framework, resulting a robust region-based recognition model. We collect a large scale dataset consisting of 138K pornographic images and 205K normal images from Internet, and our system demonstrates strong performance on this dataset, significantly outperforms traditional methods and several deep learning baselines. Additionally, the proposed method is highly efficient and real-time capable when implemented with GPU.\\

\noindent \textbf{Acknowledgements}  This work is partially supported by National Science Foundation of China (61672280,61373060,61732006), the National Key Research and Development Program of China (No. 2017YFB0802300), Jiangsu 333 Project (BRA2017377) and Qing Lan Project.

{\footnotesize
\bibliographystyle{ieee}
\bibliography{porn_recognition}
}

\begin{IEEEbiography}[{\includegraphics[width=1in,height=1.25in,clip,keepaspectratio]{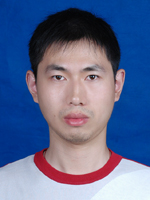}}]{\textbf{Xin Jin} received the BSc, MSc and PhD degree in the department of computer science and technology from Nanjing University of Aeronautics and Astronautics (NUAA) in 2009, 2012, and 2017. Now, he works in Megvii Research Nanjing (Face++) as a computer vision researcher.  His research interests include computer vision and deep learning, especially focusing on face landmark detection and general object detection.}
\end{IEEEbiography}

\begin{IEEEbiography}[{\includegraphics[width=1in,height=1.25in,clip,keepaspectratio]{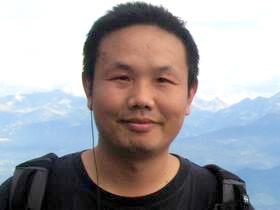}}]{\textbf{Xiaoyang Tan} received his BSc and MSc degree in computer applications from Nanjing University of Aeronautics and Astronautics (NUAA) in 1993 and 1996, respectively. Then he worked at NUAA in June 1996 as an assistant lecturer. He received a PhD degree from Department of Computer Science and Technology of Nanjing University, China, in 2005. From Sept.2006 to OCT.2007, he worked as a postdoctoral researcher in the LEAR (Learning and Recognition in Vision) team at INRIA Rhone-Alpes in Grenoble, France. His research interests are in face recognition, machine learning, pattern recognition, and computer vision. In these fields, he has authored or coauthored over 20 scientific papers.}
\end{IEEEbiography}

\end{document}